\def\year{2018}\relax
\documentclass[letterpaper]{article} 
\usepackage{aaai18}  
\usepackage{times}  
\usepackage{helvet}  
\usepackage{courier}  
\usepackage{url}  
\usepackage{graphicx}  
\newcommand*{\email}[1]{\texttt{#1}}
 
\begin{document}
%
\title{Learning User Intent from Action Sequences on Interactive Systems}
\author{Rakshit Agrawal\thanks{The first author performed the work
		while at Stubhub}\\
	University of California, Santa Cruz\\
	1156 High Street\\
	Santa Cruz, CA 95064\\
	\email{ragrawa1@ucsc.edu}
	\And
	Anwar Habeeb, Chih-Hsin Hsueh\\
	Stubhub\\
	199 Fremont Street\\
	San Francisco, CA 94105\\
	\email{\{ahabeeb,chsueh\}@stubhub.com}
}
\maketitle
\begin{abstract}
Interactive systems have taken over the web and mobile space with increasing participation from users. Applications across every marketing domain can now be accessed through mobile or web where users can directly perform certain actions and reach a desired outcome. Actions of user on a system, though, can be representative of a certain intent. Ability to learn this intent through user's actions can help draw certain insight into the behavior of users on a system.

In this paper, we present models to optimize interactive systems by learning and analyzing user intent through their actions on the system. We present a four phased model that uses time-series of interaction actions sequentially using a Long Short-Term Memory (LSTM) based sequence learning system that helps build a model for intent recognition. Our system then provides an objective specific maximization followed by analysis and contrasting methods in order to identify spaces of improvement in the interaction system. We discuss deployment scenarios for such a system and present results from evaluation on an online marketplace using user clickstream data.

\end{abstract}

\section{Introduction}
Actions, Intent, Behavior and Outcomes; all four present highly correlated characteristics of a user on an interactive system. Whether it be interaction with a search website, or a puzzle game, these four attributes always create a complex relationship of inter-dependence. Starting with an implicit \emph{behavior}, a user starts interaction with any interface, and mostly has an initial \emph{intent}. In such systems, even lack of a specific intent can be considered as being an \emph{`undecided'} intent category. While the behavior of a user has limited dependence on the system, the intent at a particular moment is still locally affected by the \emph{actions} performed during a specific period of interaction. Therefore, these actions contribute to the evolution of intent, creating a cyclic series of modified actions. The combined relation of these actions, intent and initial behavior, leads to an \emph{outcome} which can itself be either intermediate or final. In case of an intermediate outcome, it further directs actions leading up to a similar sequence again ultimately terminating at a final outcome.

On systems where users interact with web and mobile interfaces, these four attributes can both be observed and quantified up to a certain extent. For instance, on a movie review website, we can keep a record of each movie that a user has clicked on and spent time on the corresponding page. At a finer level, we can even capture actions like screen time and scroll rate on every page and each individual review. This enables us to capture data about several actions of user with exact timestamp, therefore providing a sequential data stream of actions. Depending on the kind of system, these action sequences can lead to a number of outcomes. For instance, on a marketplace website this can refer to a purchase, or a cart addition; On a reviewing platform, it can be a new review, or a comment or a `like'. The fact that these outcomes can possibly have dependence on the actions preceding them is the essential factor that we can capture through these sequences.

Ability to transform these actions into ordered sequences with available outcomes helps us use the field of supervised sequence learning in an attempt to learn models of user interaction. Since the action sequences during a session are inspired by behavior and intent, being able to learn these sequences helps us gain an insight into the underlying models that might drive these actions. Supervised sequence learning is a branch of machine learning which identifies the ordered relatedness of different data points and uses them together as an inter-linked sequential input instead of using them as independent events. Recurrent Neural Network models like Long Short-Term Memory (LSTM)~\cite{Hochreiter1997,gers_lstm} are powerful neural models that efficiently learn sequences and derive embeddings representing the implicit relationship between sequence elements. We propose using these sequential learning models to learn from action sequences on interactive systems. Such models can be trained to learn user patterns on the system corresponding to several outcomes. For example, using scroll rate and screen time along instructional videos on MOOC websites can provide a quantifiable measure of user's attentiveness towards the video. This can be further linked with potential quizzes that depend on these videos. A sequence learning model can learn the impact of these screen scrolling actions on achieved quiz score by learning scroll action sequences with the scores as a target. While these models may not be true measures of causation, they can at least learn the presence of any strong correlation between the evolving sequence patterns and outcomes.

In this paper, we present methods of deriving behavior and intent insight on web and mobile interfaces guided by tracking actions along the usage. We describe the processes for gathering potentially relevant features from actions, and representing them in the form of usable sequences. This process is followed by sequence learning on the actions to train models that correlate actions to outcomes. These trained models are then used by the system to understand potential user intent in several situations and compared against actual outcomes. Studying the predictions from these models with real evolution of a user session helps in the detection of key areas that affect change in predicted and real outcome, hence giving us a hint of actual intent. We also propose aggregation of this comparative information to identify spaces of improvement in systems targeted at a desired user outcome. 

The paper first describes components of our model in detail and explains the process from obtaining data to deriving inference. We then discuss applicability of our model in different scenarios. This is followed by experimental analysis on an online marketplace with an objective of predicting conversions. We the briefly discuss related work in this space of intent recognition and behavioral analysis. We conclude the paper with a discussion of our contributions to the field of marketing science.

\section{The Model}
User actions on an interactive system are often directed by a certain intent. Corresponding to different behaviors among users, these intents can present certain differences but due to the limitations of interface, these variations are often reasonably limited. Learning behavior of a user is a personalization property and is generally harder to learn, but intent recognition can be generalized over users and be tied instead to the interface. With the availability of usage quantification across several parameters, personalization methods have achieved high success rates in several domains. But due to the extremely large number of users and sparseness in data across parameters, learning behavior for each user still remains a challenging task. Intents, instead are more general as they have certain limitations depending on the scope of a system. While users may have many different navigation styles, the design of a website or an app can only provide a limited number of options that can be performed and therefore be tied to the intent. Therefore, this paper tries to learn user intent for a session and not specific user behavior. We expect that adding more sophisticated personalization models on our system can provide an even better understanding of user intent, but that is beyond the scope of this paper.

This paper, as described earlier focuses on using the user actions to learn intent. Actions and outcomes are the most easily attainable interactions between a user and a system. User sessions on any of the digital platforms get some kind of input from the user in the form of clicks, taps, scrolls or more complicated inputs. Any such input can be tied with a timestamp in order to make an ordered sequence of these input events. These sequences can then be represented as a function of the intent with which a user starts the concerned session. Formally, for a user $u$ during session $S$, we define the relation between actions $\alpha_u$, intent $\iota_u$, and implicit behavior $\beta_u$, as:
\begin{equation}
\label{eq_intent_action}
\alpha_u(S) = f(\iota_u(S), \beta_u, S)
\end{equation}
The dependence of actions on intent is not independent of behavior, but for learning correlations, the variation in behavior profiles might be large enough to be ignored by a learning model. With this assumption, we cluster the intents into reasonably sized groups which are much smaller than the number of behaviors observed on the system overall. Our concerned unknown in equation~\ref{eq_intent_action}, is the intent $\iota_u$. Obtaining inference on the intent directly from action sequences is not easy to achieve. The advantage of our model is the ability of our system to use action sequences as a medium to correlate intent and outcomes.

Observable actions sequences act as known variables on the system. We use the other observable quantities on the system, \emph{outcomes}, as a target for learning at the end of these action sequences. Depending on a scenario, we can measure several outcomes like purchase event, or test results, and use the action sequences to learn them. The outcome $\omega_u$ for a user session $S$, can be formally represented as:
\begin{equation}
\label{eq_action_outcome}
\omega_u(S) = g(\alpha_u(S), S)
\end{equation}

Both outcome $\omega_u$ and actions $\alpha_u$ are measurable quantities on the system and can be collected using different tracking measures. Our method first collects this dataset and then uses it to train the first stage model that depends on sequences. 

\subsection{Data Generation}
Depending on the scenario under concern, an important phase of our model is to generate structured data from the raw usage datasets for websites or apps. This data is often available in the form of raw disconnected data points. First phase of this process, therefore, is to assemble events for a session. These can be both homogeneous (eg., clicks only) or heterogeneous (eg., scrolls, clicks, taps) in nature. Session events are then filtered and ordered by their timestamps to form the action sequence. For the purpose of learning, we need to represent actions with a set of features defining them. The process of feature extraction is also specific to a system and a scenario. We perform a feature extraction process at this stage and then standardize and normalize the features in both homogeneous and heterogeneous cases respectively.

\subsection{Sequence Learning}
Second phase of our model is to perform sequence learning on the action sequences in order to learn their representation corresponding to a specific target. We use the Long Short-Term Memory (LSTM) Neural Networks for deriving embeddings corresponding to the complete action sequences. These are then sent to a sigmoid layer for deriving the final probability corresponding to the outcome. Because of sequential nature of data, LSTM ensures prediction of next event in the sequence, which is then trained to predict a specific outcome $\omega_u$. 

While the broader objective of our model is to use this trained sequential learning model for generating analysis data, the model can also be used at this stage as a prediction model. For any system, we can train multiple sequence learners for different target labels using the same set of action sequences, and use them as individual prediction models. Combined together, we can even build deeper models where the LSTMs at first layer are responsible for constructing embeddings for the action sequences and then higher layers perform predictions across different kinds of objectives.
\begin{equation}
y = \textsc{LSTM}(\alpha_u;\theta)
\end{equation}
\begin{equation}
z = \sigma(\mathbf{W}y + \mathbf{b})
\end{equation}
\begin{equation}
loss = \mathcal{L}(z, \omega_u) 
\end{equation}
where \textsc{LSTM} defines an LSTM layer with an output embedding of last activation. $\mathbf{W}$ denotes the weight matrix for sigmoid layer and $\mathbf{b}$ is the bias associated with sigmoid layer.
The $loss$ is measured using binary cross-entropy and is used to train the model using backpropagation through time.

We will denote a trained LSTM from this phase as $\textsc{LSTM}_T$. For the next phase in the model, this pre-trained $\textsc{LSTM}_T$ now acts as a function, which takes in action sequences from future dataset and generates predictions on them.

\subsection{Objective Maximization}
With the help of $\textsc{LSTM}_T$, our model now evaluates action sequences that were not a part of the training dataset, and contain true labels. We create this second dataset in order to relate the sequences with intents. Given a sequence $\alpha_u$, a real outcome $\omega_u$ and a predicted outcome $z_u$, we build a confusion matrix for the predicted intents. This provides us with sets of sequences, where predicted and real outcomes are same, and ones where the outcomes are different. For the ones with same outcomes and prediction, we do not perform any further analysis. For the sequences where outcomes vary from the prediction, we pass them through a clustering model. A specific advantage of performing clustering at this stage against clustering sequences initially is the ability to filter out significant chunks of data that can be learned using neural networks.

Since our model is aimed at improving the system for maximizing objective and not simply at predicting outcomes of sessions, we use this clustering stage as an understanding of user intent where action sequences falling within a specific cluster are assumed as belonging to a similar intent group. Each cluster is then used to analyze intent specific sessions in a detailed sequence analysis phase.

\subsection{Sequence Analysis}
After obtaining the reduced size intent clusters, we perform a detailed analysis on them that provides us with the final system specific improvement factors. Our action sequences were structured representation of user actions on the system. While $\textsc{LSTM}_T$ learned through the entire sequences, this stage evaluates each individual time step in the sequence and observes change in prediction at that stage.
For each action sequence, we generate a series of predictions $P_u$. For a sequence with length of $T$ timesteps, then we represent $P_u$ as:
\begin{equation}
P_u = \{LSTM_T(\alpha_u(t)): \forall t \in [1,T)]\}
\end{equation}
This relation can be seen as an event-wise prediction of the outcome by our sequence learner. For example, in case of a purchase outcome, and events being represented by clicks, we can consider this set as a likeliness of purchase at each click on the website.

First step in this analysis is to measure distance between predictions at each timestep. We obtain the distance set $D_u$ as:
\begin{equation}
D_u = \{dist(P_u(t)-1, P_u(t) \forall t \in [1,T)])\}
\end{equation}

$D_u$ is then sorted by the distance value. Depending on the variance across $D_u$ for different scenarios, we set a threshold value for the distances to be considered for further evaluation. By this stage we have gathered featured events $\alpha_u$ with their impact towards an outcome, along with a measured intensity of the impact using $D_u$.  We then perform final semi-automated contrasting between feature vectors, sequences and predictions in order to explore interface events that create a higher distance between prediction and reality.

\subsection{Semi-Automated Contrasting}
This is the final phase of our model which is currently performed semi-automatically by an expert of the system. The sorted impact sequences using $D_u$ provide us with action events causing drastic changes in prediction. We combine sequences with such features together and observe overall impact caused by them on the predictions. In cases of strong significance, we are able to identify features of the system that can be potential causes of the change. A set of such evaluated features is then used to improve the system for maximizing specific objectives.

\begin{figure}
	\includegraphics[height=1.5in, width=3in]{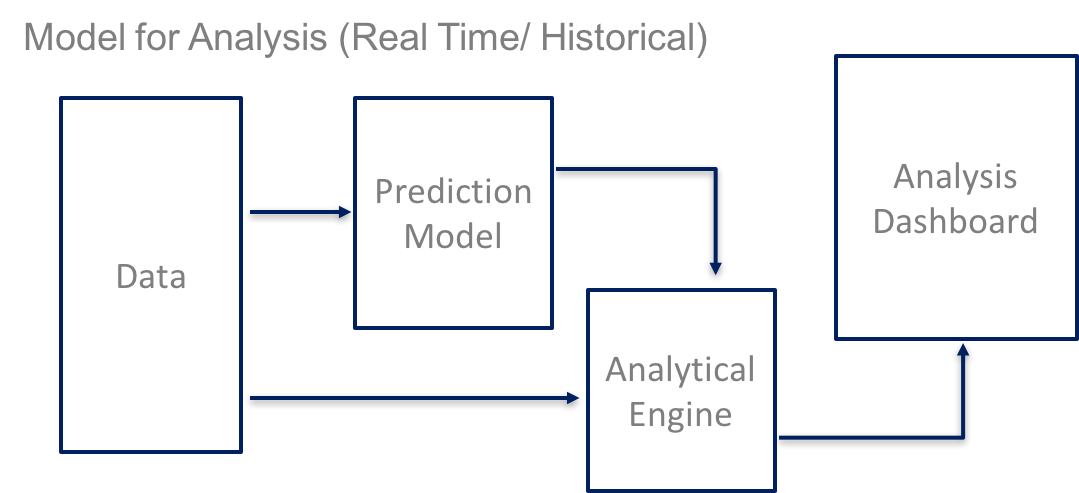}
	\caption{Model structure for implementing the stages till semi-automated contrasting}
	\label{fig_analysis_dashboard}
\end{figure}

\begin{figure}
	\includegraphics[height=1.5in, width=3in]{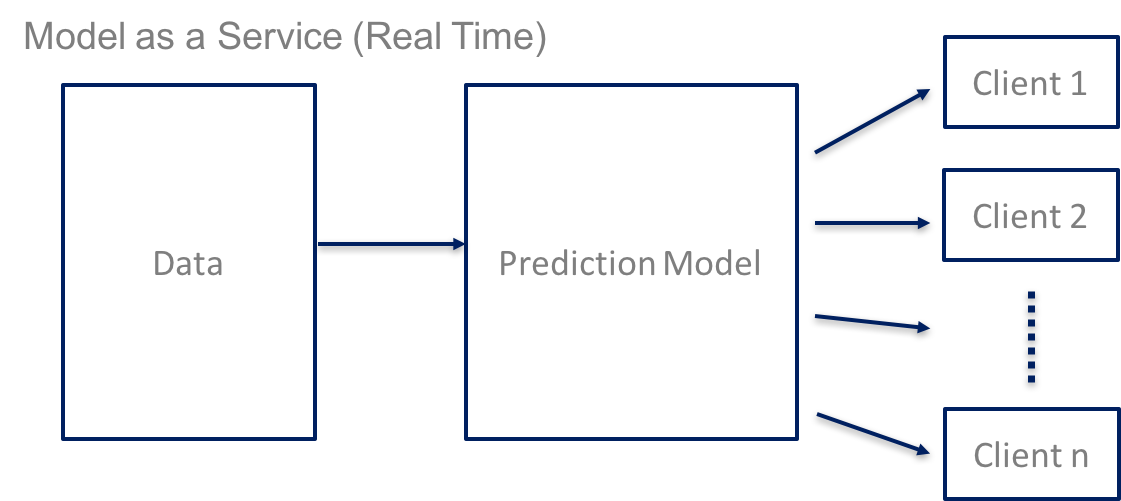}
	\caption{Model as a service for predicting outcome, to be used differently}
	\label{fig_model_as_service}
\end{figure}

Our complete system is a combination of these modules that allow for condition-specific learning in any interactive framework. Combined together these modules can be used to explore usage across a specific objective.
We also propose the usage of this model in active and passive form. As a passive system, it can be used to analyze sessions through a dashboard similar to the architecture shown in Figure~\ref{fig_analysis_dashboard}. A complete analysis model consists of the sequence learning, analysis, and contrasting modules. Modularity of our system also allows for an active usage of the system as a prediction service model during a session. Since the sequence learning model can provide predictions at each time step, it can be used in real time with any client for providing intent prediction. Figure~\ref{fig_model_as_service} depicts the architecture for using this system as a service with multiple clients.

\section{Use Case Scenarios}
In the previous section, we described how our model operates at each stage. In this section we discuss practical scenarios and systems where this model can be used. We also discuss some approaches to be followed after the semi-automated contrasting using our model to maximize concerned objectives.

In general, our model can be used in any interactive system where user provides connected inputs at different points in time during a session on the platform. Since our model only requires the two observable quantities, action sequences and outcomes, we focus mainly on systems which can provide action sequences with certain features. We also assume that each system has at least one objective function which is relevant to it in some way and whose maximization can benefit the system.

\subsection{Online Marketplaces}
Our first scenario is for online marketplaces in the form of websites and apps. Online marketplaces cover the wide range of websites where some form of purchase can be made from a larger set of items. These can include shopping websites, event ticketing websites, and more such platform where user purchase is a desired outcome. Conversion rate is one important metric of such systems, and therefore, can be a significant objective for our system to maximize. Several more outcomes, like adding items to cart, returning a purchase, selling an item, etc can also be studied in such systems. Actions on such platforms comprise of clicks across different components like items, pages, categories, filters among others. Actions can also include scrolling events, viewing, zooming and more interaction functions provided by the application.

Using our model, we first convert user interactions with the marketplace into an ordered sequences. We then derive features along each event. This can correspond to features like time distance between clicks, category corresponding to the clicked item, type of page where click was performed, and more specific details. All these features can potentially add relevant information to the overall model. We then train the Sequence Learning module to learn a model on the action sequences for predicting the outcome. This process is then followed by the remaining phases of our model to correlate events causing major change in the prediction between different timesteps. 

\subsection{Online Coursewares}
Online coursewares are another significant form of interactive systems where users watch videos, read content, perform quizzes among other variations. A general objective of coursewares is to ensure learning among the users and to be able to distribute content in the best possible way. The objective function, in these cases, therefore, tries to maximize the learning outcome. This can be measured using scores on the quizzes, often provided at the end of video or interactive learning sessions. 

Session in this scenario can capture screen time spent by the user on videos, scroll rate during reading content or while watching the video, clicks or highlights in the reading content, amount of answer switching on quizzes, and more specific practices. All together, these sessions can provide a time-series of actions with rich meta-data along with a wide evaluated range of targets to learn. Targets can be scores on final quizzes, responses on course surveys, or some system specific measures.

Using our model, similar to marketplaces, we will capture ordered sequence of the input actions, and will extract features for each action event. We will then perform Sequential Learning on these actions with the model objective of predicting a measure of user's performance in the course. 

\subsection{Other scenarios}
While we discussed two specific cases, our proposed approach is applicable in a much broader variety of scenarios. Capturing meta data along user sessions in the form of timed sequences along with system specific targets can mostly be used to improve system. For instance, on specific interest based websites like cooking, biking, or arts, objectives are around improving readership and promoting discussions on posts. Similarly forums provide discussion channels that can be targeted at increasing responses or answers for emerging questions. Our model can be similarly applied to such systems, by using meta-data across usage and learning the patterns of usage directed at maximizing desired response. We do not go into details of these scenarios as the breadth in their range is wide and the paper is focused on the structured method for learning user intent, and not necessarily on the use cases of learned intent.

\section{Model Evaluation}

\begin{figure*}
	\includegraphics[width=0.33\linewidth]{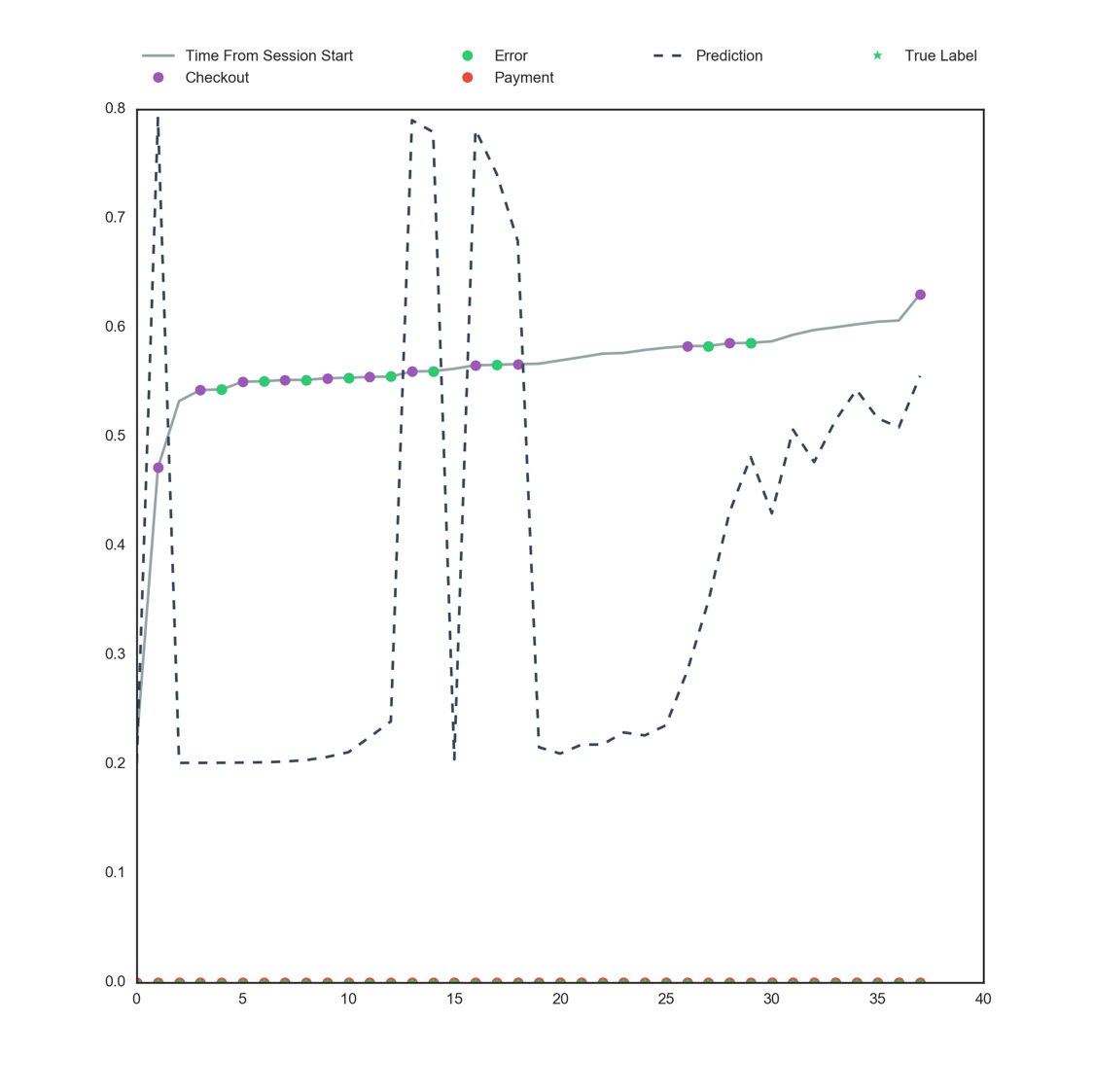}
	\includegraphics[width=0.33\linewidth]{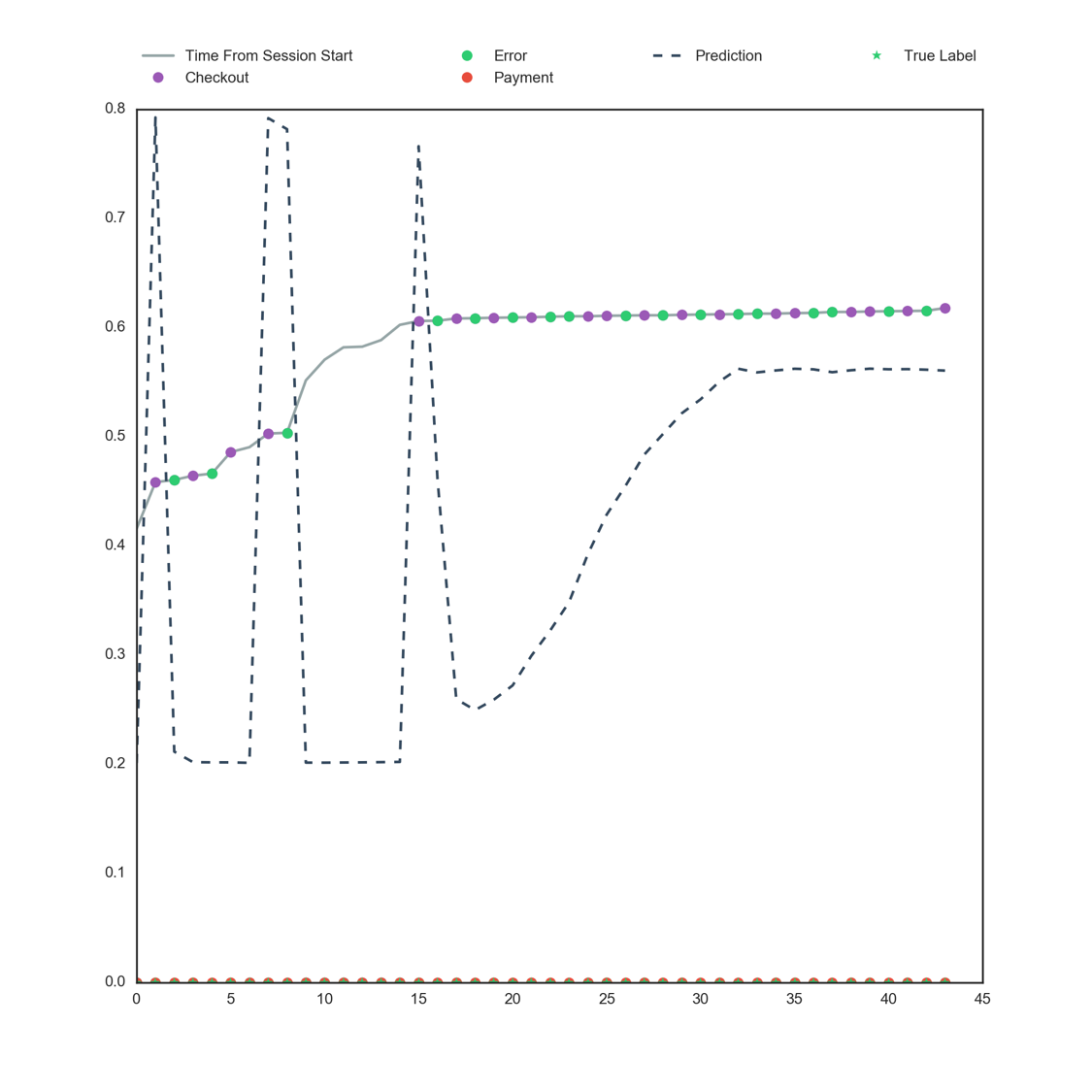}
	\includegraphics[width=0.33\linewidth]{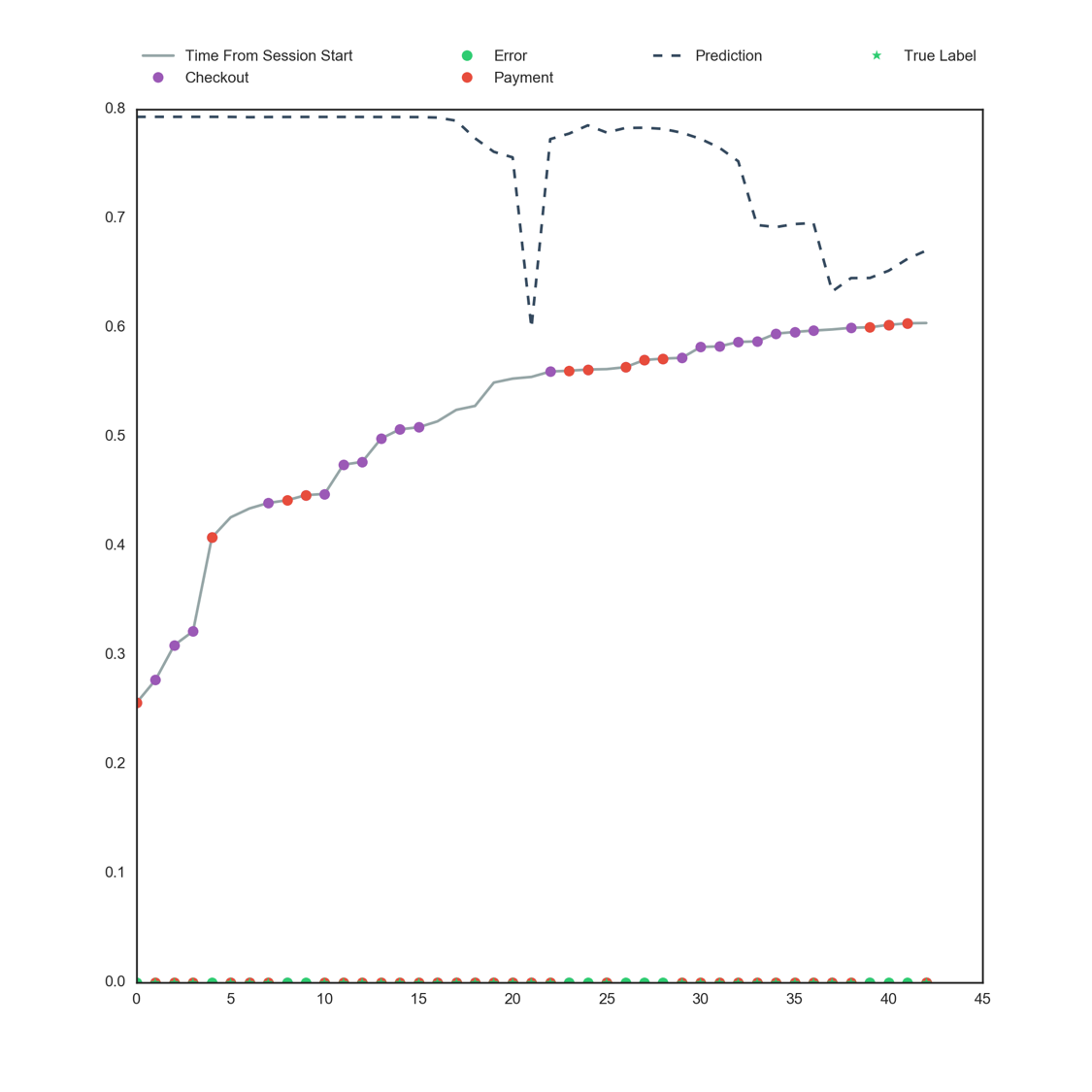}
	\caption{Plots of samples from click sessions with prediction for \emph{`purchase'} along the session with events at each click}
	\label{fig_graph_result}
\end{figure*}

We experimented and evaluated our model for the scenario of online marketplaces. Our data was collected from a ticketing website where users can \emph{sell} or \emph{purchase} tickets. Data was completely anonymized and each session was independent of any user specific parameters. We considered each user session as a unique entity and captured the action events for each click on the website during that session. This provided us with a time series of clicks for each session, where properties of these click were used to derive features within the sequence.

For generating our training data, we sampled sessions from each hour of the day over three months in late 2016. Our neural model consisting of LSTM and feedforward neural network layers was built on Keras~\cite{keras} with Tensorflow~\cite{Tensorflow} backend. Target label for our dataset was the presence or absence of a \emph{conversion} in the session. We trained the model using binary cross-entropy as our loss function, and used the Adadelta~\cite{adadelta} optimizer.
We evaluated performance of the trained model on sessions from both past and future months outside the training window. Our analysis data consisted of sessions in months from early 2016, and early 2017. 

We evaluated the result of the neural model, signifying the probability of a \emph{conversion within a session} with real labels from data. This system, when evaluated on 1-click before the final outcome achieved an average accuracy of $0.89$, and an average recall o $0.85$.
We also evaluated the neural model against varying number of steps, $k$, before outcome. We observed a significant monotonic improvement in the prediction as we got closer to the end point. Average recall recorded by the model on test data at $k=4$, was $0.70$, whereas at $k=2$, it was $0.81$. At $k=1$, the penultimate step, the recall rises to an average of $0.85$.
This consistent improvement in the ability of our model to predict conversions also shows its capability to use and improve with additional local action context in order to predict outcome.

While a high accuracy strengthens the reliability of model alignment with user intent prediction, another attribute of our system is to use this knowledge for improving the system. This was performed by the sequential analysis process where we used trained learning model to visualize variations in predicted intent along with actions of the user. Any significant change in prediction over subsequent clicks was then used to evaluate feature change during those clicks. We derived graphical representations of clicks across some of the page actions like `Checkout', `Error', etc. Plots showing the change in prediction with clicks over time are presented in figure~\ref{fig_graph_result}.

\section{Related Work}
In this paper we presented models for intent recognition on interactive systems. The space of web interactions has been studied in several fields including Computer Science, Psychology and Economics for behavioral analysis and intent recognition.
\cite{Radinsky:2012:MPB:2187836.2187918} explores behavior on web systems and makes use of this information in order to predict system parameters. 
\cite{Benevenuto:2009:CUB:1644893.1644900} makes use of the actions performed on a web system by collecting clickstream data and tries to derive inference based on those topics.

\cite{Obendorf2007WebPR} performed a study on browser usage by using click-streams which are similar to our action sequences. This study made use of the similar sequences across different websites in order to understand parameters for the browser. We perform evaluation on the application under concern and treat the web browser as an independent platform.
\cite{Wu:2009:PCP:1645953.1646127} used linear models for predicting conversion likeliness among users with the help of usage features derived over time.
Recently,~\cite{Sun:2017:UCC:3063955.3063971} presented analysis on courseware clickstream data in order to improve systems.
\cite{eye_shopping} use eye tracking methods for identifying intent of shopping. They study the dependence on eye actions directed towards system's outcome.
Another common approach of defining user behavior on e-commerce websites is by building neural models as defined in~\cite{Borisov2016ANC} and~\cite{Wu2015NeuralMO}. 

While these works present different ways to analyze usage data directed at certain objective, our work provides a more concrete model of relating action sequences with intent recognition with the use of sequence learning.
Sequence learning as a field has gained enormous success over past few years with LSTMs particularly being used in several cases.
\cite{Graves2008} detailed the use of LSTM in speech recognition, followed by variants like bi-directional LSTMs~\cite{Graves2013} and Grid LSTMs~\cite{Kalchbrenner2015}.

\section{Conclusions}
In this paper, we presented a structured model with multiple modules aimed at improving interactive systems on web and mobile platforms. The paper presented strong relations between \emph{intent} and \emph{actions} and formalized the process of intent recognition. 
We presented a novel method of relating the sequences derived from different action events on a system, and learning their representations for further inference. By using session outcomes as a label for supervised sequence learning, we presented models that can efficiently predict future actions, giving hints on the intent of user.

Our model also described active and passive learning systems, where an active system can use these predictions in real time, and a passive system, which can use the learning model to analyze usage on the system. We also presented models for analysis and contrasting that can help detect system characteristics that affect user's actions during a session.

We have successfully deployed this system on an online ticketing marketplace and have discussed its potential use cases where this exact model can be used to optimize other system objectives. The model, in general, is independent of the nature of system and can be deployed in any setting which captures user actions. The time-series nature of this model makes it a great architecture for exploration with different sequence learning methods depending on nature of data.

Beyond intent recognition, the paper also identifies relation between user behavior and actions. While we use presented models to obtain inference only on the intent and not on the behavior, we believe our models can benefit future studies in the space of behavior modeling as well. With the help of observed actions and outcomes, we can use a trained model to capture intent along different sessions of different users. We can further identify more specific attributes of users and sessions in order to relate intent with the underlying behavior.
Modular structure of our model also allows for additional information to be used at any stage of the model. For instance, the sequential learning module provides us with a representation of next predicted action. If a system obtains more useful features after certain clicks, this representation can be used along with additional features in order to draw final prediction.

Through the presented models, we hope to provide assistance in various application spaces, and expect for research in the space of marketing science to improve with clearer understanding of user intent.

\bibliographystyle{aaai}
\bibliography{aaai18}

\end{document}